\documentclass[lettersize,journal]{IEEEtran}
\usepackage{amsmath,amsfonts}
\usepackage{algorithmic}
\usepackage{algorithm}
\usepackage{array}
\usepackage[caption=false,font=normalsize,labelfont=sf,textfont=sf]{subfig}
\usepackage{textcomp}
\usepackage{stfloats}
\usepackage{url}
\usepackage{verbatim}
\usepackage{graphicx}
\usepackage{cite}
\usepackage{listings}
\usepackage{tikz}
\newcommand{\circled}[1]{\tikz[baseline=(char.base)]{
  \node[shape=circle,draw,inner sep=1pt] (char) {#1};}}

\usepackage{authblk}

\usepackage{hyperref}
\usepackage{enumitem}
\usepackage{booktabs} 
\usepackage[table]{xcolor}
\usepackage{colortbl}     
\usepackage{multirow} 
\usepackage{amsmath}
\usepackage{makecell}
\usepackage{caption}
\usepackage{tabularx, array}
\usepackage{xspace}

\usepackage[most]{tcolorbox}

\usepackage{wrapfig}

\newcommand{\revised}[1]{{\color{black} #1}}

\newcommand{\boxmargin}{2mm}
\tcbuselibrary{most, skins, breakable, theorems}
\newtcolorbox{myboxa}[2][]{
    colback=gray!10!white,
    colframe=black, enhanced,
    attach boxed title to top left={yshift=-2mm,xshift=5mm},
    title=#2,#1
}
\newtcolorbox{myboxb}[2][]{
    boxsep=3pt,
    left = \boxmargin, right = \boxmargin, top = \boxmargin, bottom = \boxmargin,
    title={#2},#1
}
\newtcolorbox{myboxc}{
    colback=yellow!15!white,
    colframe=gray!,
    arc = 0pt, outer arc = 5pt,
    boxsep=1pt,
    leftrule=3pt,
    bottomrule=0pt, toprule=0pt, rightrule=0pt,
    left = \boxmargin, right = \boxmargin, top = \boxmargin, bottom = \boxmargin,
    before skip=5pt,
    after skip=5pt
}
\newtcolorbox{myboxd}{
    colback=gray!10,
    colframe=black,
    width=\columnwidth,
    arc=1mm, auto outer arc,
    boxrule=0.5pt,
}

\usepackage{tikz}

\begin{document}

\title{PhoenixRepair: Rethinking Repair Strategy Exploration in Software Agents}

\author[1]{Tianyue Jiang}
\author[1,†]{Yanlin Wang \thanks{† Yanlin Wang is the corresponding author.}}
\author[1]{Xin He}
\author[1]{Daya Guo}
\author[2]{Jiachi Chen}
\author[3]{Ming Wen}
\author[4]{Ensheng Shi}
\author[4]{Xilin Liu}
\author[4]{Yuchi Ma}
\author[1]{Guanbin Li}
\affil[1]{Sun Yat-sen University, China}
\affil[2]{State Key Laboratory of Blockchain and Data Security, Zhejiang University, China}
\affil[3]{School of Cyber Science and Engineering, Huazhong University of Science and Technology, China}
\affil[4]{Huawei CodeArts Model Team, China}

\markboth{IEEE Transactions on Software Engineering,~Vol.~XX, No.~X, XX~2026}%
{Jiang \MakeLowercase{\textit{et al.}}: PhoenixRepair: Expanding Software Agents' Search Space and Refining Solutions through Self-Reflection}


\maketitle

\begin{abstract}
While Large Language Models have greatly advanced automated issue resolution, existing agent-based methods exhibit a fundamental limitation in their insufficient exploration of repair strategies. This insufficiency manifests in two key aspects. First, the exploration of multiple potential edit locations is limited. Second, the exploration of repair attempts at each location is also insufficient. To address these challenges, we present PhoenixRepair, a multi-agent framework that systematically explores multiple candidate edit locations and performs iterative reflection and refinement on patch generation, thereby expanding the search space of repair strategies. Our framework begins with multi-location sampling, optionally augmented with graph-based localization information for difficult tasks, followed by iterative reflection and refinement to generate better patches, culminating in final-round generation guided by distilled insights from all historical attempts. Experiments on SWE-bench-Verified demonstrate that PhoenixRepair achieves the largest relative improvement of 7.8\% over SWE-agent under DeepSeek-V3.1, and attains the highest resolved rate of 76.0\% Pass@1 under MiniMax-M2.5. Meanwhile, it achieves higher fault localization accuracy than existing approaches. Our code is available at \url{https://github.com/DeepSoftwareAnalytics/PhoenixRepair}.
\end{abstract}


\IEEEdisplaynontitleabstractindextext

\IEEEpeerreviewmaketitle

\section{Introduction}
\IEEEPARstart{I}{ssue} resolution is an important part of modern software development and is crucial for ensuring software system quality~\cite{zhang2024autocoderover}. The automating in this process improves development efficiency and ultimately contributes to the sustainability of software projects. With strong code comprehension and reasoning capabilities, Large Language Models (LLMs) have greatly advanced this field~\cite{liu2024deepseek, achiam2023gpt, yang2025qwen3}.
To measure LLMs' ability to handle authentic software engineering problems, Jimenez et al. introduced SWE-bench, a benchmark built upon real GitHub issues~\cite{jimenez2023swe}.
Recently, considerable research has focused on developing agent-based methods to solve the tasks in SWE-bench~\cite{yang2024swe, gao2025trae, wang2024openhands}. Such agent-based methods typically equip LLMs with predefined tools, enabling LLMs to autonomously execute actions, observe feedback, and plan subsequent steps~\cite{liu2024largesurvey}. These tools may include functionalities such as creating/writing files, running tests, and more.

\begin{figure}[t]
    \centering
    \includegraphics[width=1\columnwidth]{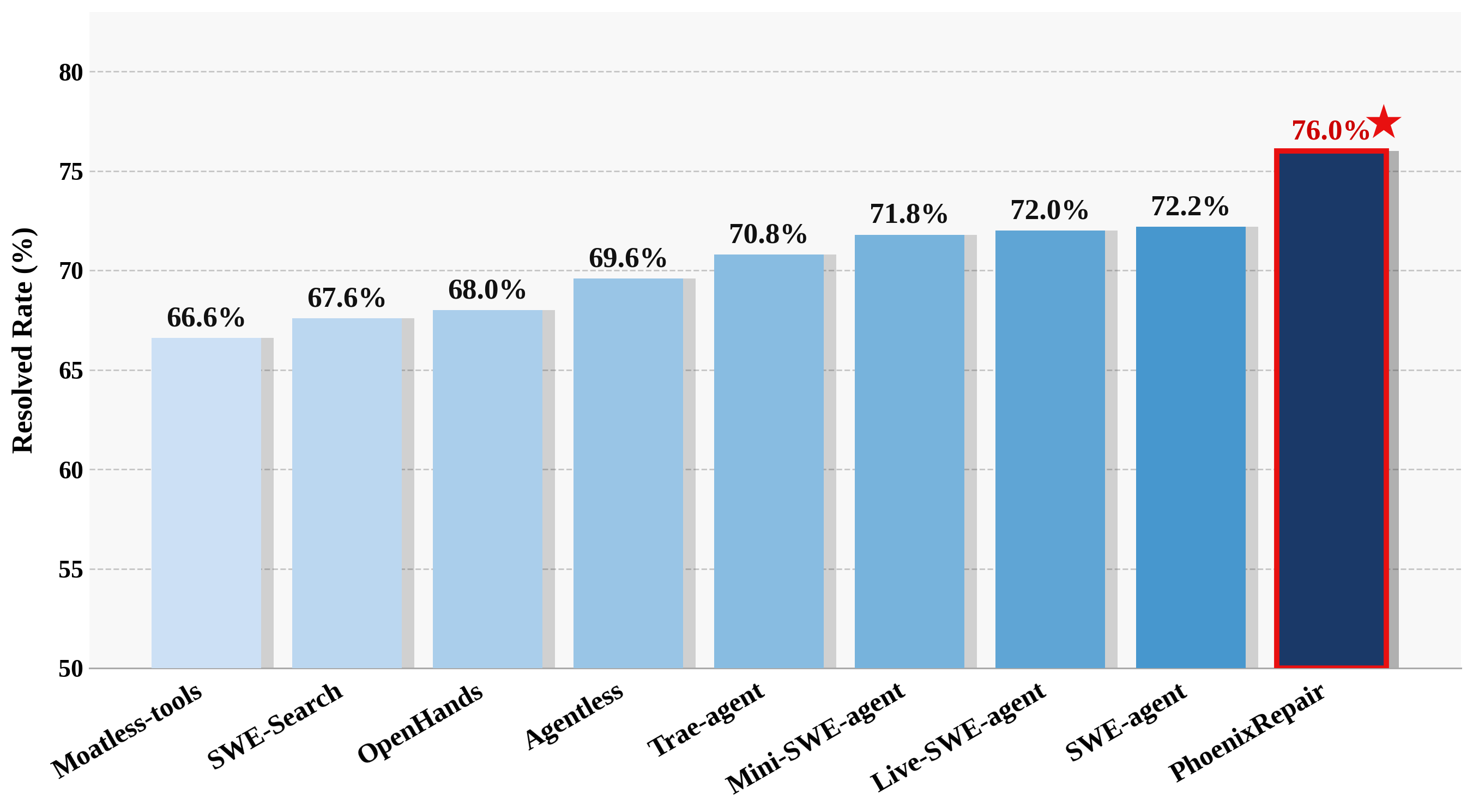}
    \caption{Comparison of MiniMax-M2.5 Performance Under Different Frameworks on SWE-Bench-Verified.}
    \label{fig:a}
\end{figure}

\begin{figure*}[t]
    \centering
    \includegraphics[width=0.9\textwidth]{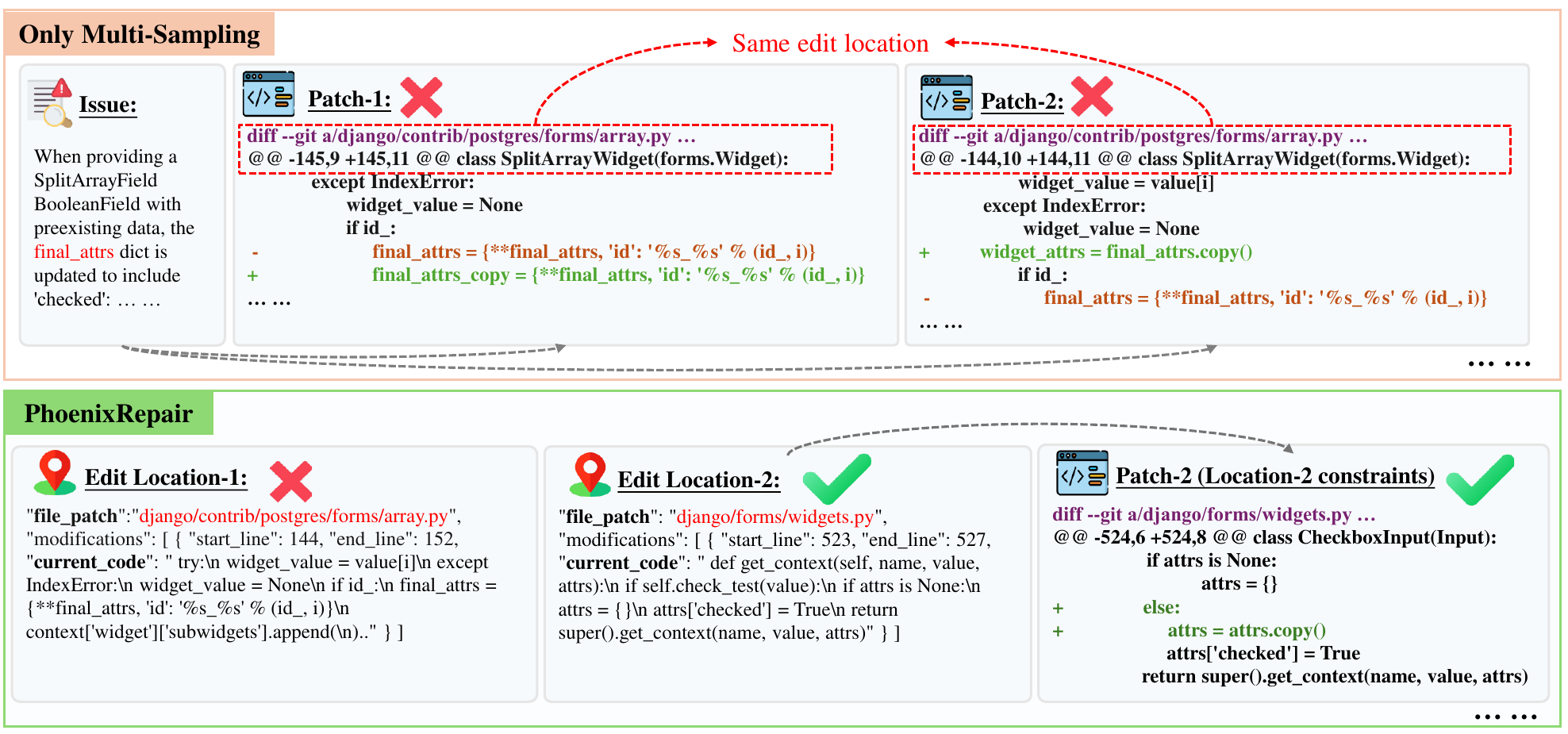}
    \caption{A motivating example of solving \texttt{django\#12193}, the agent consistently attempts repairs at an incorrect location across multiple sampling, but generates the right patch when constrained to the correct location.}
    \label{fig:motivation}
\end{figure*}

A key challenge in issue resolution lies in fault localization~\cite{guo2016deep}. The complexity arises from the fact that multiple code snippets may be relevant to a given issue, resulting in multiple candidate repair strategies. However, existing agent-based methods suffer from insufficient exploration of repair strategies, which may lead them to select suboptimal repair strategies and apply repairs directly. Specifically, this insufficient exploration appears in two aspects:

\textbf{P1: The exploration of multiple potential edit locations is insufficient.} Simply employing test-time scaling strategies~\cite{ehrlich2025codemonkeys} to sample execution trajectories multiple times (e.g., by varying system prompts or temperature) fails to fully address this limitation. As shown in Figure~\ref{fig:motivation}, simply running multiple samples does not guarantee coverage of different edit locations. When solving \texttt{Django\#12193}, the agent assigns high confidence to \texttt{forms/array.py} and repeatedly samples it across five parallel runs, leading all repair attempts to concentrate on an incorrect location. This is attributed to the inherent characteristics of probabilistic language models, which exhibit a bias toward high-probability solution patterns, thereby restricting the diversity of the search space~\cite{kalavasis2025limits, linguisticdecline2024, alignment2025}. SE-Agent~\cite{lin2025se} first samples multiple trajectories to build a trajectory pool. It then mutates or recombines the trajectories in the pool to further expand the search space. However, if all sampled trajectories focus on a confident but incorrect location, trajectory-level mutation or recombination become ineffective in producing a correct repair.

\textbf{P2: The exploration of repair attempts on each edit location is insufficient.} As shown in Figure~\ref{fig:casestudy}, a single repair attempt at the correct edit location may still be insufficient, as the generated patch can remain problematic. In the case, the first repair at the correct location \emph{location-1} incompletely implemented the ignore logic. It failed to exclude the current directory itself before the \texttt{yield} statement. As a result, ignored directories could still be yielded and linted. In the second round of repair at \emph{location-1}, the omission was addressed by adding an explicit ignore check prior to yielding. This change fully prevented ignored paths from participating in recursive linting. This example highlights the importance of iterative reflection and refinement at each edit location.

To address these limitations, we propose \textbf{PhoenixRepair}, a multi-agent framework that explores multiple candidate edit locations, which effectively \textbf{addresses P1}. It further performs iterative reflection and refinement on patches generated for each edit location, thereby \textbf{addressing P2}. 
Our framework employs a localization sampling agent that iteratively samples edit locations. For tasks with more difficult fault localization, we utilize a graph-based localization agent to provide additional cross-file dependency information. For each sampled edit location, a coder agent generates candidate patches, while a test and selection agent selects the top-performing ones based on multiple indicators. An analysis agent distills guidance from historical attempts at each selected edit location. The coder agents then use this guidance to regenerate patches at these locations. This process continues until convergence to a final round edit location, at which point the analysis agent distills guidance from all previous attempts on the final edit location to direct the patch generation in the final round.

Our experimental evaluation demonstrates the effectiveness of PhoenixRepair, which consistently achieves the highest Pass@1 across all evaluated backbone model. Our evaluation demonstrates the effectiveness of PhoenixRepair, which consistently achieves the highest Pass@1 across all evaluated models, attaining up to 76.0\% with MiniMax-M2.5 and improving fault localization accuracy across all granularities. We further validate its generalizability by integrating the proposed mechanisms into Mini-SWE-agent and Live-SWE-agent, where consistent improvements are observed. Overall, these results confirm the effectiveness, stability, and transferability of PhoenixRepair for automated issue resolution. Our main contributions are summarized as follows:

\begin{itemize}
    \item We propose PhoenixRepair, a multi-agent framework that addresses the insufficient exploration in repair strategies exhibited by existing agent-based approaches through multiple edit locations sampling and iterative reflection and refinement mechanism.
    \item Our experiments demonstrate that our framework outperforms other state-of-the-art methods under the same model configuration. Specifically, PhoenixRepair achieves the largest relative improvement of 7.8\% over SWE-agent under DeepSeek-V3.1, and attains the highest resolved rate of 76.0\% Pass@1 under MiniMax-M2.5.
    \item Our proposed mechanisms generalize well across different agent-based methods, such as Mini-SWE-agent and Live-SWE-agent, validating the broad applicability of our approach.
\end{itemize}

\begin{figure*}[t]
    \centering
    \includegraphics[width=0.9\textwidth]{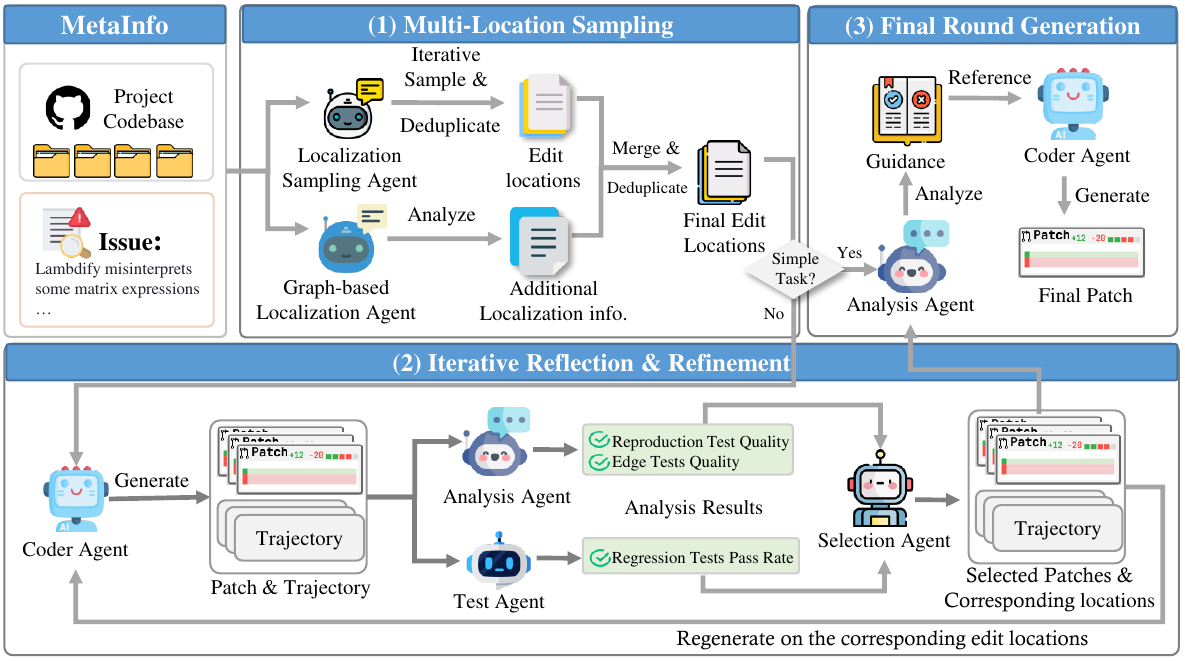}
    \caption{Overview of our PhoenixRepair framework. The framework comprises three main phases: (1) multiple location sampling, (2) iterative reflection and refinement mechanism, and (3) final round generation.}
    \label{fig:overview}
\end{figure*}

\section{Methods}

Existing agent-based methods for issue resolution often suffer from insufficient exploration of the search space of repair strategies. To address this problem, we propose PhoenixRepair, a multi-agent framework that increases exploration in two ways. First, it samples multiple candidate edit locations instead of committing to a single one. Second, for each candidate location, it runs iterative reflection and refinement to improve the patch across rounds, using feedback from past attempts.

As shown in Figure~\ref{fig:overview}, PhoenixRepair proceeds in three phases: \circled{1} \textbf{Multiple location sampling.} A localization agent executes several localization sampling and outputs a set of candidate edit locations. For hard tasks, we further add graph-based localization results to expand this set, and then deduplicate the locations. \circled{2} \textbf{Iterative reflection and refinement.} A coder agent generates one patch for each candidate location by reproducing the issue, editing the code at the specified location, and validating the fix. A selector agent then ranks all candidates using test quality and regression test pass rate, keeps the top half, and repeats this generate--analyze--select loop until only one edit location remains. \circled{3} \textbf{Final-round generation.} Finally, an analysis agent summarizes the full history of attempts at the remaining edit location and produces guidance. The coder agent then generates the final patch following this guidance.

\subsection{Multi-location Sampling}

Given the issue description and repository context, a localization
sampling agent performs $N$ iterations of code localization, yielding
$N$ candidate edit locations. These iterations are executed
sequentially: in the $i$-th iteration (where $1 < i \leq N$), the
agent is provided with the previously obtained edit locations
$\{\ell_1, \ell_2, \ldots, \ell_{i-1}\}$ as references to explore
alternative candidates, thereby expanding the search space. Each edit
location $\ell_j$ is specified by: (1) the file path $f$, (2) the
start and end line numbers $(l_s, l_e)$, and (3) the target code
snippet for modification $c$. We perform a deduplication on the $N$ edit locations obtained from multiple sampling iterations. 
Following prior study~\cite{xia2024agentless}, two edit locations are considered duplicates if they share the same file path, correspond to the same program entity (e.g., the same method within a class), and exhibit overlapping line ranges. After removing such redundant locations, we obtain a deduplicated edit locations, denoted as $\mathcal{L}'$, where $|\mathcal{L}'| \leq N$. We categorize all tasks based on the number of locations in the deduplicated edit locations $\mathcal{L}'$. Specifically, we divide tasks into three categories according to the difficulty of fault localization. Tasks with relatively simple fault localization satisfy $|\mathcal{L}'| = 1$. Tasks with medium fault localization difficulty satisfy $1 < |\mathcal{L}'| \leq k$. Tasks with more challenging fault localization satisfy $|\mathcal{L}'| > k$.


Tasks with more challenging fault localization retain more than $k$ edit locations after deduplication ($|\mathcal{L}'| > k$). This reflects increased uncertainty with respect to identifying the correct edit location for repair. For these difficult tasks, we incorporate additional localization information from a graph-based localization agent~\cite{chen2025locagent}, which considers cross-file dependency information during fault localization. 
This agent constructs a heterogeneous graph representation of the repository, capturing code structures and dependencies such as
import relationships, function invocations, and class inheritance
hierarchies. 
Based on the issue description and the
dependency graph, the agent identifies relevant code entities (e.g.,
\texttt{django/forms/widgets.py:\allowbreak CheckboxInput.get\_context}),
which may not be explicitly referenced in the issue but are connected
through multi-hop dependencies.

To extract the top $l$ ranked entities, the agent generates an initial ranking and refines it through cross-iteration confidence estimation, selecting entities that consistently rank higher across iterations and ultimately choosing the top $l$ entities from the sorted list. Then we treats each entity along with its associated metadata (file path, class name, function name, and line numbers) as an individual edit location. We denote these graph-based edit locations as 
and refer to them as additional localization information. These edit locations are then merged with those obtained from our localization sampling agent: $\mathcal{L}'' = \mathcal{L}' \cup \mathcal{L}_{\text{graph}}$. After this augmentation, we perform deduplication using the same criteria described above to obtain the final edit locations $\mathcal{L}_{\text{final}}$ for subsequent patch generation.

The final edit locations used for patch generation differ across task categories. For tasks with simple or medium fault localization difficulty, corresponding to $|\mathcal{L}'| = 1$ or $1 < |\mathcal{L}'| \leq k$, we directly set the final edit location set as $\mathcal{L}_{\text{final}} = \mathcal{L}'$. For tasks with challenging fault localization, where $|\mathcal{L}'| > k$, we augment the deduplicated locations with additional localization information, yielding $\mathcal{L}'' = \mathcal{L}' \cup \mathcal{L}_{\text{graph}}$. After applying deduplication, the resulting set $\mathcal{L}_{\text{final}}$ is used as the final candidate edit locations for patch generation, where $\mathcal{L}_{\text{final}} \subseteq \mathcal{L}''$.

\subsection{Iterative Reflection And Refinement}
In the previous stage, tasks are categorized into three groups based on the difficulty of fault localization. Tasks with simple or medium fault localization result in a final set of edit locations obtained directly from the deduplicated localization results. However, tasks with challenging fault localization incorporate additional graph-based localization information before deduplication. 
For tasks with medium or challenging fault localization, we constrain a coder agent through the system prompt to generate a separate repair patch for each of the $w$ locations in $\mathcal{L}_{\text{final}}$, yielding $w$ corresponding execution trajectories and associated patches. 
Then we employ a selection agent to identify the top $\lceil w/2 \rceil$ patches along with their corresponding edit locations. The selection agent performs this selection by valuating each candidate patch and its associated execution trajectory. To ensure a reliable and well-rounded assessment, the selection agent adopts a multi-dimensional quality evaluation framework that consists of two indicators:

\paragraph{Reproduction \& Edge Tests Quality}

We denote the Reproduction \& Edge Test Quality as $Q_{\text{test}}(\tau_i)$. The value $Q_{\text{test}}(\tau_i)$ is performed by an analysis agent, which evaluates the comprehensiveness and quality of both reproduction tests and edge tests generated throughout the execution trajectory $\tau_i$. Before performing the assessment, we preprocess the execution trajectory by compressing it to remove redundant information.
Specifically, the compression step eliminates repetitive actions, intermediate failed attempts. Actions that repeatedly trigger identical error messages or execute unchanged test scripts are merged into a single representative step.  
After compression, the trajectory retains the essential test construction steps, key execution outcomes, and validated test cases. The analysis agent then evaluates the quality of the reproduction tests and edge tests based on this compact representation, enabling a more accurate and noise-resistant assessment of test coverage and effectiveness.

\paragraph{Regression Test Pass Rate}

We measure the regression test pass rate $R_{\text{pass}}(p_i)$ to evaluate the impact of a candidate patch $p_i$ on the existing codebase, specifically to assess whether applying the patch introduces unintended side effects or new errors in previously correct functionality. This procedure follows a three-stage pipeline adopted in prior work~\cite{gao2025trae, xia2024agentless}. In the first stage, we extract all existing test cases from the original codebase and execute them on the original source code. All tests that pass are collected and used to construct the initial regression test set $\mathcal{T}_{\text{init}}$. In the second stage, a test agent further filters the initial regression test set to identify a subset of tests that are most likely to represent behaviors that should remain unchanged after the fix. Specifically, the agent selects tests that exercise functionality unrelated to the reported issue and are expected to continue passing under a correct patch. This filtering step is necessary because not all passing tests are suitable as regression tests. In particular, fixing certain issues may legitimately alter existing behavior, which can cause some previously passing tests to fail. As a result, such tests should not be treated as indicators of unintended regressions. Through this process, we obtain the final regression test set, denoted as $\mathcal{T}_{\text{final}}$. In the third stage, we perform patch validation. Each candidate patch $p_i$ is applied to the original codebase and evaluated on the final regression test set $\mathcal{T}_{\text{final}}$. We compute the regression pass rate as the proportion of tests in $\mathcal{T}_{\text{final}}$ that pass after applying $p_i$.

A selection agent then ranks all $w$ patches using a weighted combination of the reproduction \& edge test quality $Q_{\text{test}}(\tau_i)$ and the regression test pass rate $R_{\text{pass}}(p_i)$. It selects the top $\lceil w/2 \rceil$ patches and their corresponding edit locations.

\subsection{Final-round Generation}

In the final round, for tasks with medium or difficult fault localization, the analysis agent distills insights from all previous execution trajectories and their corresponding patches generated at the final round edit location $\ell_{\text{final}}$. Based on this historical information, the agent produces guidance that directs the final patch generation. 
For tasks with simple fault localization, we have already constrained the coder agent to generate patches at the edit position specified by the final round edit location $\ell_{\text{final}}$, sampling $t$ patches in parallel. We then employ the analysis agent to distill these $t$ patches and their corresponding execution trajectories, producing guidance information.
Once the guidance information has been distilled, we inject it into the system prompt to provide contextual direction for the patch generation. By leveraging the distilled insights, the actor agent can generate patches that gain experience from historical attempts and are also aligned with the specified edit locations.

\section{Experiments}

In our experiments, we investigate the following research questions to thoroughly evaluate the effectiveness and performance of PhoenixRepair:

\begin{itemize}
    \item \textbf{RQ1:} What is the overall resolution rate of PhoenixRepair, and how does it compare to the other baselines?
    \item \textbf{RQ2:} How does PhoenixRepair perform in fault localization when compared to other baseline methods?
    \item \textbf{RQ3:} To what extent do the mechanisms proposed in PhoenixRepair generalize to other agent-based methods, and can they be seamlessly integrated as plug-in components to further enhance their performance?
    \item \textbf{RQ4:} What is the contribution of each individual component within our framework, and how do they collectively impact the overall performance of PhoenixRepair?
    \item \textbf{RQ5:} How much inference cost does PhoenixRepair consume compared to baselines approaches?
\end{itemize}

\subsection{Experimental Settings}
\subsubsection{\textbf{Datasets}} We conduct experiments with PhoenixRepair on SWE-Bench-Verified~\cite{openai2024introducing}, which contains 500 manually verified issues from SWE-bench~\cite{jimenez2023swe}. 

\subsubsection{\textbf{Metrics}}

We employ the following metrics to evaluate the performance of all frameworks:

\begin{itemize}[left=4pt]
    \item \textbf{Pass@1:} Following previous studies~\cite{jimenez2023swe, yang2024swe, antoniades2024swe}, we evaluate model performance using Pass@1, which measures the proportion of issues correctly resolved by the first generated patch. 

\item \textbf{Acc@1:}
Following prior work~\cite{xia2024agentless, chen2025locagent, antoniades2024swe}, we evaluate the accuracy of localization at three different levels of granularity.

\begin{itemize}
    \item \textit{File level:} We report the metric \textbf{File Acc@1}. A prediction is considered correct if the edit location contains all files that require modification.
    \item \textit{Module or class level:} We report the metric \textbf{Module Acc@1}. A prediction is considered correct if the edit location contains all classes or modules that require modification.
    \item \textit{Function or method level:} We report the metric \textbf{Function Acc@1}. A prediction is considered correct if the edit location contains all functions and methods that require modification.
\end{itemize}
\end{itemize}




\subsubsection{\textbf{Baselines}}

We compare PhoenixRepair with two categories of competitive baselines:

\textbf{Procedure-based methods.} We compare with \textbf{Agentless}~\cite{xia2024agentless}, a non-agentic pipeline that decomposes the repair process into several phases of localization, repair, and patch validation.

\textbf{Agent-based methods.} We include several competitive agent-based frameworks specifically designed for automated program repair.

\begin{itemize}[left=4pt]
    \item \textbf{SWE-Search~\cite{antoniades2024swe}:} This method integrates Monte Carlo Tree Search (MCTS) to systematically explore the repair space~\cite{antoniades2024swe}.
    \item \textbf{OpenHands~\cite{wang2024openhands}:} This approach provides a sandboxed execution environment that supports code editing and test execution, and incorporates agents such as CodeActAgent to evaluate candidate repairs~\cite{wang2024openhands}.
    \item \textbf{SWE-agent~\cite{yang2024swe}:} This system introduces an Agent-Computer Interface that enables agents to navigate repositories, edit files, and run commands end-to-end to resolve issues~\cite{yang2024swe}.
    \item \textbf{Mini-SWE-agent~\cite{minisweagent2024}:} This is a minimalist agent scaffold that retains strong performance~\cite{minisweagent2024}.
    \item \textbf{Live-SWE-agent~\cite{xia2025live}:} This agent autonomously evolves its agent scaffold during runtime to better solve the current issue~\cite{xia2025live}.
    \item \textbf{Trae-agent~\cite{gao2025trae}:} This method adopts an agent-based approach with execution feedback and testing signals to improve patch generation and selection~\cite{gao2025trae}.
\end{itemize}

\subsubsection{\textbf{Implementation Details}}
We implement PhoenixRepair by extending SWE-agent~\cite{yang2024swe} with our multi-location sampling and iterative reflection and refinement mechanisms. The graph-based localization agent is based on the LocAgent~\cite{chen2025locagent}. In our experiments, we evaluate PhoenixRepair across five backbone models: DeepSeek-V3.1, DeepSeek-V3.2, Qwen-Coder-Plus, GLM-4.7, and MiniMax-M2.5. For each task, we sample edit locations for five iterations. A task is classified as having high fault localization difficulty if the number of deduplicated edit locations exceeds four. For tasks with difficult fault localization, we retain the top two entities identified by the graph-based localization agent. For tasks with simple fault localization, we perform three parallel executions on the final round edit location.

\subsection{RQ1: Overall Performance}
 In RQ1, we investigate whether expanding the repair strategy search space through our proposed mechanisms can improve issue resolution performance. Table~\ref{tab:overall} summarizes the Pass@1 of all methods. Overall, PhoenixRepair achieves the best performance under each model configuration, demonstrating its effectiveness in issue resolution. Among the agent-based methods, SWE-agent serves as the strongest baseline, yet PhoenixRepair surpasses it in every experimental setting. Under DeepSeek-V3.1, PhoenixRepair achieves a relative gain of 7.8\%, the largest among all five settings, suggesting that the proposed mechanisms are especially effective when the underlying model has moderate capability and thus greater room for improvement through systematic strategy exploration. Under Qwen-Coder-Plus and GLM-4.7, PhoenixRepair yields a consistent relative improvement of 6.4\% in both cases, further supporting the robustness of the approach across different models. Under DeepSeek-V3.2, PhoenixRepair attains a Pass@1 of 74.4\%, representing a relative improvement of 7.2\% over SWE-agent. Even with MiniMax-M2.5, which yields the highest absolute resolved rate of 76.0\% among all configurations, PhoenixRepair still improves over SWE-agent by 5.3\% relatively, demonstrating that the benefits of expanded strategy search persist even as the backbone model becomes more capable. The consistent gains across all five backbones indicate that the improvements are not attributable to a specific model, but rather to the proposed multi-location sampling and iterative reflection and refinement mechanisms, which systematically increase the diversity and quality of candidate repair strategies explored during inference.

\begin{table*}[!t]
    \centering
    \small
    \setlength{\tabcolsep}{12pt}
    \renewcommand{\arraystretch}{0.95}
    \caption{Comparison of overall performance with baseline approaches. Each cell shows the resolved rate (\%) and number of resolved tasks (\#). Red percentages indicate relative performance gains over SWE-agent.}
    \label{tab:overall}
    \begin{tabular}{lccccc}
        \toprule
        \textbf{Method}
            & \textbf{DeepSeek-V3.1}
            & \textbf{Qwen-Coder-Plus}
            & \textbf{DeepSeek-V3.2}
            & \textbf{GLM-4.7}
            & \textbf{MiniMax-M2.5} \\
        \cmidrule(lr){2-6}
        & \multicolumn{5}{c}{\textit{Resolved Rate (\%) (Resolved Tasks \#)}} \\
        \midrule
        \multicolumn{6}{c}{\textbf{Procedure-based methods}} \\
        \midrule
        Agentless
            & 59.0 (295) & 63.8 (319) & 67.0 (335) & 62.6 (313) & 69.6 (348) \\
        \midrule
        \multicolumn{6}{c}{\textbf{Agent-based methods}} \\
        \midrule
        Moatless-tools
            & 55.6 (278) & 60.0 (300) & 64.8 (324) & 59.8 (299) & 66.6 (333) \\
        Mini-SWE-agent
            & 55.4 (277) & 59.8 (299) & 61.0 (305) & 64.8 (324) & 71.8 (359) \\
        Live-SWE-agent
            & 56.2 (281) & 60.2 (301) & 63.8 (319) & 64.8 (324) & 72.0 (360) \\
        SWE-Search
            & 57.6 (288) & 61.2 (306) & 65.4 (327) & 61.2 (306) & 67.6 (338) \\
        OpenHands
            & 57.2 (286) & 60.6 (303) & 62.8 (314) & 60.2 (301) & 68.0 (340) \\
        Trae-agent
            & 59.4 (297) & 64.6 (323) & 67.8 (339) & 64.4 (322) & 70.8 (354) \\
        SWE-agent
            & 61.2 (306) & 65.4 (327) & 69.4 (347) & 65.6 (328) & 72.2 (361) \\
        \midrule
        \multicolumn{6}{c}{\textbf{Our method}} \\
        \midrule
        \rowcolor{cyan!20}
        \textsc{PhoenixRepair}
            & 66.0 (330)$^{\textcolor{red}{\uparrow7.8\%}}$
            & 69.6 (348)$^{\textcolor{red}{\uparrow6.4\%}}$
            & 74.4 (372)$^{\textcolor{red}{\uparrow7.2\%}}$
            & 69.8 (349)$^{\textcolor{red}{\uparrow6.4\%}}$
            & 76.0 (380)$^{\textcolor{red}{\uparrow5.3\%}}$ \\
        \bottomrule
    \end{tabular}
\end{table*}

Compared with Agentless, PhoenixRepair achieves consistent and substantial relative improvements across all model configurations, ranging from 9.1\% to 11.9\%. These results highlight the inherent limitation of fixed-stage repair pipelines, which commit to a repair strategy without the ability to revise it based on runtime execution signals. By contrast, PhoenixRepair integrates repair generation with iterative execution feedback, enabling dynamic strategy adjustment that proves critical for resolving complex issues. 

These results confirm that explicitly expanding the repair strategy search space through multi-location exploration and iterative refinement is an effective and model-agnostic approach for improving the performance of automated issue resolution, and underscore the importance of systematic exploration beyond single-location or single-attempt repair strategies.

\begin{center}
    \begin{myboxc}
    \textbf{RQ1 Summary: }
PhoenixRepair consistently achieves the highest Pass@1 across all five backbone models, validating that multi-location sampling and iterative reflection and refinement yield substantial, model-agnostic improvements over both agent-based and procedure-based baselines.
    \end{myboxc}
\end{center}

\begin{table}[!t]
\centering
\small
\caption{Localization accuracy comparison results. Red percentages denote PhoenixRepair's relative improvements over SWE-agent. We evaluate methods with two LLMs: DeepSeek-V3.2 (DS-V3.2) and Qwen3-Coder-Plus (Qwen3-CP).}
\label{tab:location}
\setlength{\tabcolsep}{2pt}
\renewcommand{\arraystretch}{1.0}
\begin{tabular}{lllll}
\toprule
\textbf{Method} & \textbf{LLM} &
\makecell[l]{\textbf{File}\\\textbf{Acc@1}} &
\makecell[l]{\textbf{Module}\\\textbf{Acc@1}} &
\makecell[l]{\textbf{Function}\\\textbf{Acc@1}} \\
\midrule
OpenHands        & DS-V3.2 & 79.64 & 69.58 & 60.82 \\
SWE-Search       & DS-V3.2 & 81.62 & 73.98 & 63.06 \\
Trae-agent       & DS-V3.2 & 82.84 & 75.38 & 64.82 \\
Mini-SWE-agent   & DS-V3.2 & 79.84 & 71.42 & 61.84 \\
Live-SWE-agent   & DS-V3.2 & 81.24 & 73.42 & 62.84 \\
\midrule
\multirow{2}{*}{SWE-agent}        & Qwen3-CP & 79.62 & 71.98 & 61.24 \\
 & DS-V3.2    & 83.04 & 74.22 & 63.42 \\
\midrule
\multirow{2}{*}{PhoenixRepair}    & Qwen3-CP
& 82.24$^{\textcolor{red}{\uparrow 3.29\%}}$
& 74.22$^{\textcolor{red}{\uparrow 3.11\%}}$
& 64.42$^{\textcolor{red}{\uparrow 5.19\%}}$ \\
 & DS-V3.2
& \textbf{85.64}$^{\textcolor{red}{\uparrow 3.13\%}}$
& \textbf{76.82}$^{\textcolor{red}{\uparrow 3.50\%}}$
& \textbf{66.64}$^{\textcolor{red}{\uparrow 5.08\%}}$ \\
\bottomrule
\end{tabular}
\end{table}

\subsection{RQ2: Fault Localization Performance}

Table~\ref{tab:location} presents a comparison of fault localization accuracy across three granularities, including file-level, module-level, and function-level Acc@1. Overall, PhoenixRepair consistently achieves the highest localization accuracy among all evaluated methods at each granularity, demonstrating its effectiveness in improving fault localization performance.

Under the DeepSeek-V3.2 configuration, PhoenixRepair attains Acc@1 scores of 85.64\%, 76.82\%, and 66.64\% at the file, module, and function levels, respectively, yielding relative improvements of 3.13\%, 3.50\%, and 5.08\% over SWE-agent across the three granularities. Similar trends are observed under the Qwen3-Coder-Plus configuration, where PhoenixRepair achieves file-level, module-level, and function-level Acc@1 scores of 82.24\%, 74.22\%, and 64.42\%, corresponding to relative improvements of 3.29\%, 3.11\%, and 5.19\% over SWE-agent. Notably, improvements are more pronounced at the function level across both configurations, indicating that PhoenixRepair is particularly effective at localizing faults to fine-grained code regions. While other agent-based baselines achieve competitive localization performance, their accuracy remains consistently lower than that of PhoenixRepair, especially at the function level. These consistent gains across all granularities suggest that expanding the exploration of candidate edit locations and incorporating an iterative refinement mechanism jointly contribute to more accurate fault localization outcomes, not only at the coarse file level but also at finer-grained code scopes.

\begin{center}
    \begin{myboxc}
    \textbf{RQ2 Summary: }
PhoenixRepair achieves consistent improvements in fault localization accuracy across file, module, and function levels, with particular gains at the function level indicating its ability to narrow the search space to precise code regions and benefit subsequent repair generation.
    \end{myboxc}
\end{center}

\subsection{RQ3: Generalizability}

To validate the generalizability of our method, we integrate our multi-location sampling and iterative reflection and refinement mechanisms into the Mini-SWE-agent and Live-SWE-agent. The experimental settings are identical to those used when extending SWE-agent, including all associated parameter configurations.
All experiments in this section use DeepSeek-V3.2 as the backbone LLM. As shown in Table~\ref{tab:transfer}, methods annotated with \emph{w/ ReflLoc} indicate the corresponding agent augmented with our multi-location sampling and iterative reflection and refinement mechanisms.

For Mini-SWE-agent, incorporating our mechanisms leads to consistent relative improvements across all evaluation metrics. Specifically, the file-level, module-level, and function-level Acc@1 scores increase relatively by 3.06\%, 3.16\%, and 2.72\%, respectively. In addition, the overall Pass@1 score improves from 61.0\% to 65.2\%, corresponding to a relative gain of 6.9\%. These results indicate that the proposed mechanisms enhance both fault localization accuracy and end-to-end issue resolution performance when applied to the Mini-SWE-agent.

\begin{table}[!t]
\centering
\small
\caption{Performance comparison when applying our mechanism to other agent-based methods.}
\label{tab:transfer}
\setlength{\tabcolsep}{2pt}
\renewcommand{\arraystretch}{1.2}
\begin{tabular}{lllll}
\toprule
\textbf{Method}
& \makecell[l]{\textbf{File}\\\textbf{Acc@1}}
& \makecell[l]{\textbf{Module}\\\textbf{Acc@1}}
& \makecell[l]{\textbf{Function}\\\textbf{Acc@1}}
& \textbf{Pass@1} \\
\midrule
Mini-SWE-agent
& 79.84 & 71.42 & 61.84 & 61.0 \\
\ \textit{w/ ReflLoc}
& 82.28$^{\textcolor{blue}{\uparrow 3.06\%}}$
& 73.68$^{\textcolor{blue}{\uparrow 3.16\%}}$
& 63.52$^{\textcolor{blue}{\uparrow 2.72\%}}$
& 65.2$^{\textcolor{red}{\uparrow 6.9\%}}$ \\
\midrule
Live-SWE-agent
& 81.24 & 73.42 & 62.84 & 63.8 \\
\ \textit{w/ ReflLoc}
& 83.76$^{\textcolor{blue}{\uparrow 3.10\%}}$
& 75.64$^{\textcolor{blue}{\uparrow 3.02\%}}$
& 64.88$^{\textcolor{blue}{\uparrow 3.25\%}}$
& 67.4$^{\textcolor{red}{\uparrow 5.6\%}}$ \\
\bottomrule
\end{tabular}
\end{table}

Similar trends are observed for Live-SWE-agent. With our mechanisms enabled, file-level, module-level, and function-level Acc@1 scores increase relatively by 3.10\%, 3.02\%, and 3.25\%, respectively, while the Pass@1 score improves from 63.8\% to 67.4\%, corresponding to a relative gain of 5.6\%. The consistent improvements across different granularities and both agent-based frameworks demonstrate that expanding candidate edit location exploration and incorporating iterative refinement positively affect localization accuracy and ultimately contribute to improved resolution performance. These gains further indicate that our proposed mechanisms can be seamlessly integrated as plug-in components into different agent architectures, yielding consistent performance improvements and demonstrating robust generalizability.


\begin{center}
    \begin{myboxc}
    \textbf{RQ3 Summary: }
On both Mini-SWE-Agent and Live-SWE-Agent, our multi-location sampling and iterative reflection and refinement mechanisms yield consistent gains in Pass@1 and fault localization accuracy, indicating robust generalization to different agent-based methods.
    \end{myboxc}
\end{center}

\subsection{RQ4: Ablation Study}

We validate the effectiveness of each component of PhoenixRepair through the ablation study. In the study, ``w/o Additional Location Info.'' refers to removing the additional localization information provided by the graph-based localization agent. ``w/o Multi-Location Sampling'' denotes applying the iterative reflection and refinement mechanism to a single sampled edit location instead of iteratively sampling multiple edit locations. ``w/o Iterative Refl.'' indicates a configuration in which the iterative reflection and refinement mechanism is disabled for each edit location, eliminating iterative feedback during the repair process.
Based on Table~\ref{tab:ablation}, we observe that removing any component leads to a degradation in Pass@1. Under the Qwen3-Coder-Plus, removing the additional location information results in a relative decrease of 0.3\%, whereas removing multi-location sampling or iterative reflection leads to larger relative drops of 2.9\% and 3.4\%. A similar pattern is observed under the DeepSeek-V3.2.
These results indicate that all components contribute positively to overall performance, with multi-location sampling and iterative reflection having a larger relative impact on performance.

\begin{figure}[!t]
    \centering
    \includegraphics[width=0.95\columnwidth]{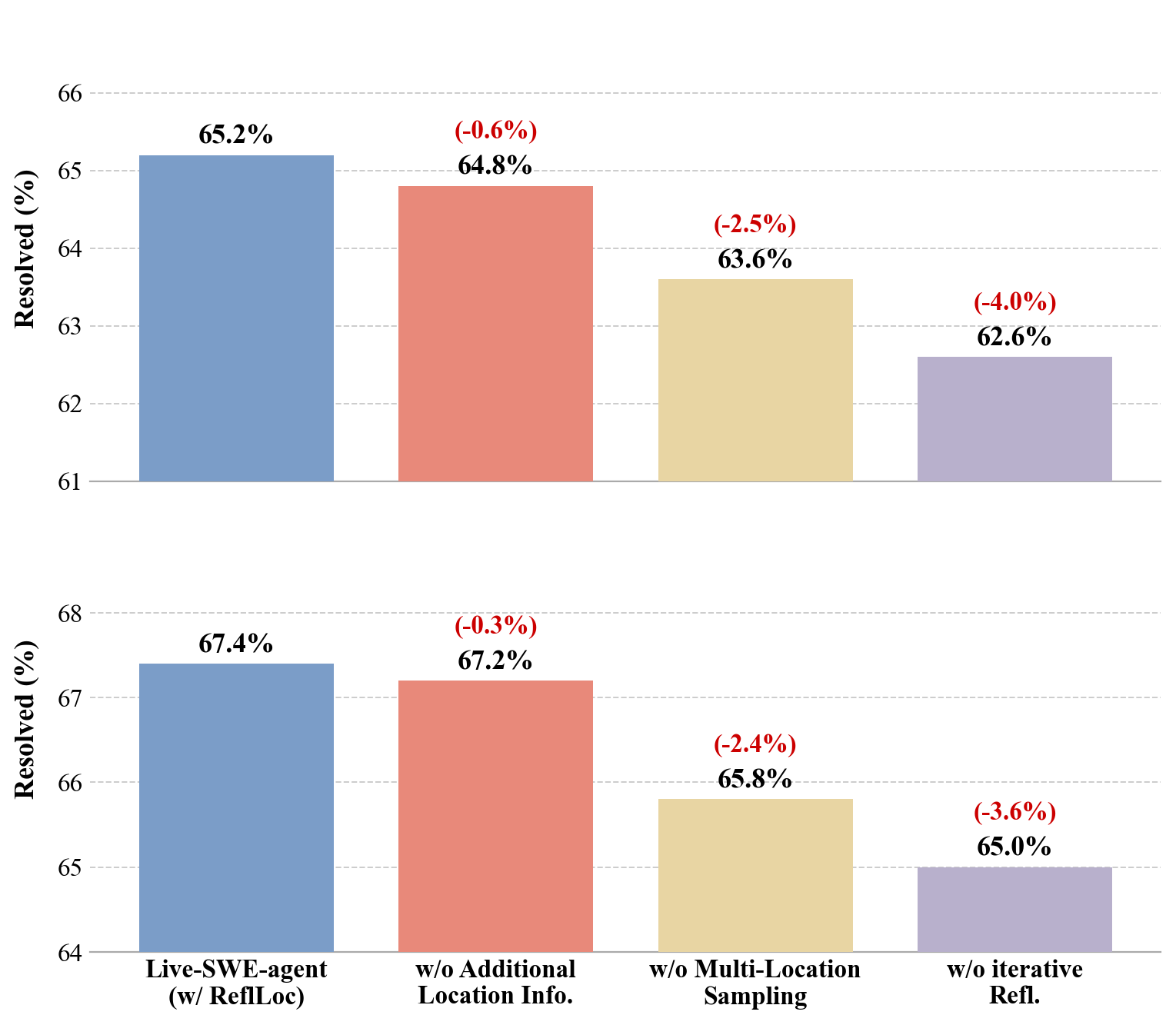}
    \caption{Ablation study results for Mini-SWE-agent (w/ ReflLoc) and Live-SWE-agent (w/ ReflLoc).}
    \label{fig:ablation}
\end{figure}

Figure~\ref{fig:ablation} further reports ablation results when applying our mechanisms to Mini-SWE-agent (w/ ReflLoc) and Live-SWE-agent (w/ ReflLoc). The methods annotated with \emph{w/ ReflLoc} indicate the corresponding agent augmented with multi-location sampling and iterative reflection and refinement mechanisms. For Mini-SWE-agent (w/ ReflLoc), removing additional location information leads to a modest relative decrease of 0.6\% in resolved rate, while removing multi-location sampling and iterative reflection results in larger relative degradations of 2.5\% and 4.0\%, respectively. Similarly, for Live-SWE-agent (w/ ReflLoc), removing additional location information causes a relative drop of 0.3\%, whereas removing multi-location sampling and iterative reflection leads to relative decreases of 2.4\% and 3.6\%, respectively. The relative magnitudes of these drops are consistent with the trends observed in Table~\ref{tab:ablation}.

\begin{table}[!t]
\centering
\small
\caption{Ablation study results with different backbone models. $\Delta$ denotes the relative performance change compared to the full PhoenixRepair system.}
\label{tab:ablation}
\setlength{\tabcolsep}{1pt}
\begin{tabular}{lcccc}
\toprule
\multirow{2}{*}{\textbf{Method}}
& \multicolumn{2}{c}{\textbf{Qwen3-Coder-Plus}}
& \multicolumn{2}{c}{\textbf{DeepSeek-V3.2}} \\
\cmidrule(lr){2-3} \cmidrule(lr){4-5}
& \textbf{Pass@1} & \textbf{$\Delta$(\%)}
& \textbf{Pass@1} & \textbf{$\Delta$(\%)} \\
\midrule
PhoenixRepair
& 69.6\% & --
& 74.4\% & -- \\
\ \ w/o Additional Location Info.
& 69.4\% & \textcolor{red}{$\downarrow$0.3}
& 74.2\% & \textcolor{red}{$\downarrow$0.3} \\
\ \ w/o Multi-Location Sampling
& 67.6\% & \textcolor{red}{$\downarrow$2.9}
& 72.2\% & \textcolor{red}{$\downarrow$3.0} \\
\ \ w/o Iterative Refl.
& 67.2\% & \textcolor{red}{$\downarrow$3.4}
& 71.4\% & \textcolor{red}{$\downarrow$4.0} \\
\bottomrule
\end{tabular}
\end{table}

\begin{center}
    \begin{myboxc}
    \textbf{RQ4 Summary: }
The results of the ablation study validate the effectiveness of our two core mechanisms, namely multi-location sampling and iterative reflection, confirming that each component contributes meaningfully to the overall performance of PhoenixRepair.
    \end{myboxc}
\end{center}

\subsection{RQ5: Cost Analysis}
To address concerns regarding the computational overhead introduced by PhoenixRepair's multi-location sampling and iterative reflection and refinement mechanisms, we conduct a detailed cost analysis comparing PhoenixRepair against baseline methods under the DeepSeek-V3.2.

\subsubsection{Raw API Cost}

Table~\ref{tab:cost} presents the average API cost per task for each method. PhoenixRepair incurs an average cost of \$0.154 per task, compared to \$0.083 for SWE-agent, \$0.079 for Mini-SWE-agent, and \$0.080 for Live-SWE-agent. This corresponds to approximately a 1.85$\times$ overhead over the SWE-agent. We consider this overhead reasonable given the consistent and substantial improvements in Pass@1 and fault localization accuracy.

\subsubsection{Prefix Caching Reduces Effective Overhead}
Modern LLM inference providers offer prefix-caching mechanisms that substantially reduce the cost of repeated context. For instance, DeepSeek and Anthropic cache identical prompt prefixes across API calls, reducing the cost of cached tokens by up to 90\%~\cite{deepseek2024contextcaching, anthropic2024promptcaching}. PhoenixRepair benefits from prefix caching in two primary ways:

\begin{itemize}
    \item \textbf{Shared system prompt and repository context across sampling iterations.} During multi-location sampling, the localization agent performs $N$ iterations of code localization over the same repository context and issue description. Since the system prompt and repository context remain identical across these iterations, the corresponding tokens are nearly always served from cache, substantially reducing the marginal cost of each additional sampling iteration.
    \item \textbf{Shared repository context and task description across branch conversations.} During the iterative reflection and refinement phase, the coder agent generates patches independently for each candidate edit location. Each of these branch conversations shares the same repository context and task description. Prefix caching ensures that this shared context is loaded only once, reducing the effective token cost across all parallel branches.
\end{itemize}

\begin{table}[!t]
\centering
\small
\caption{Average API cost per task under DeepSeek-V3.2. Overhead is computed relative to SWE-agent.}
\label{tab:cost}
\setlength{\tabcolsep}{2pt}
\renewcommand{\arraystretch}{1.1}
\begin{tabular}{lccc}
\toprule
\textbf{Method} & \textbf{Avg. Cost / Task (\$)} & \textbf{Pass@1 (\%)} & \textbf{Overhead} \\
\midrule
Mini-SWE-agent   & 0.079 & 61.0 & 0.95$\times$ \\
Live-SWE-agent   & 0.080 & 63.8 & 0.96$\times$ \\
SWE-agent        & 0.083 & 69.4 & 1.00$\times$ \\
\midrule
\textsc{PhoenixRepair} & \textbf{0.154} & \textbf{74.4} & 1.85$\times$ \\
\bottomrule
\end{tabular}
\end{table}

\begin{center}
    \begin{myboxc}
    \textbf{RQ5 Summary: }
Owing to prefix caching from modern LLM providers and our engineering optimizations, PhoenixRepair's effective cost in practice is considerably lower than raw API counts suggest, achieving a favorable balance between performance gains and inference overhead.
    \end{myboxc}
\end{center}

\begin{figure*}[!t]
    \centering
    \includegraphics[width=0.85\textwidth]{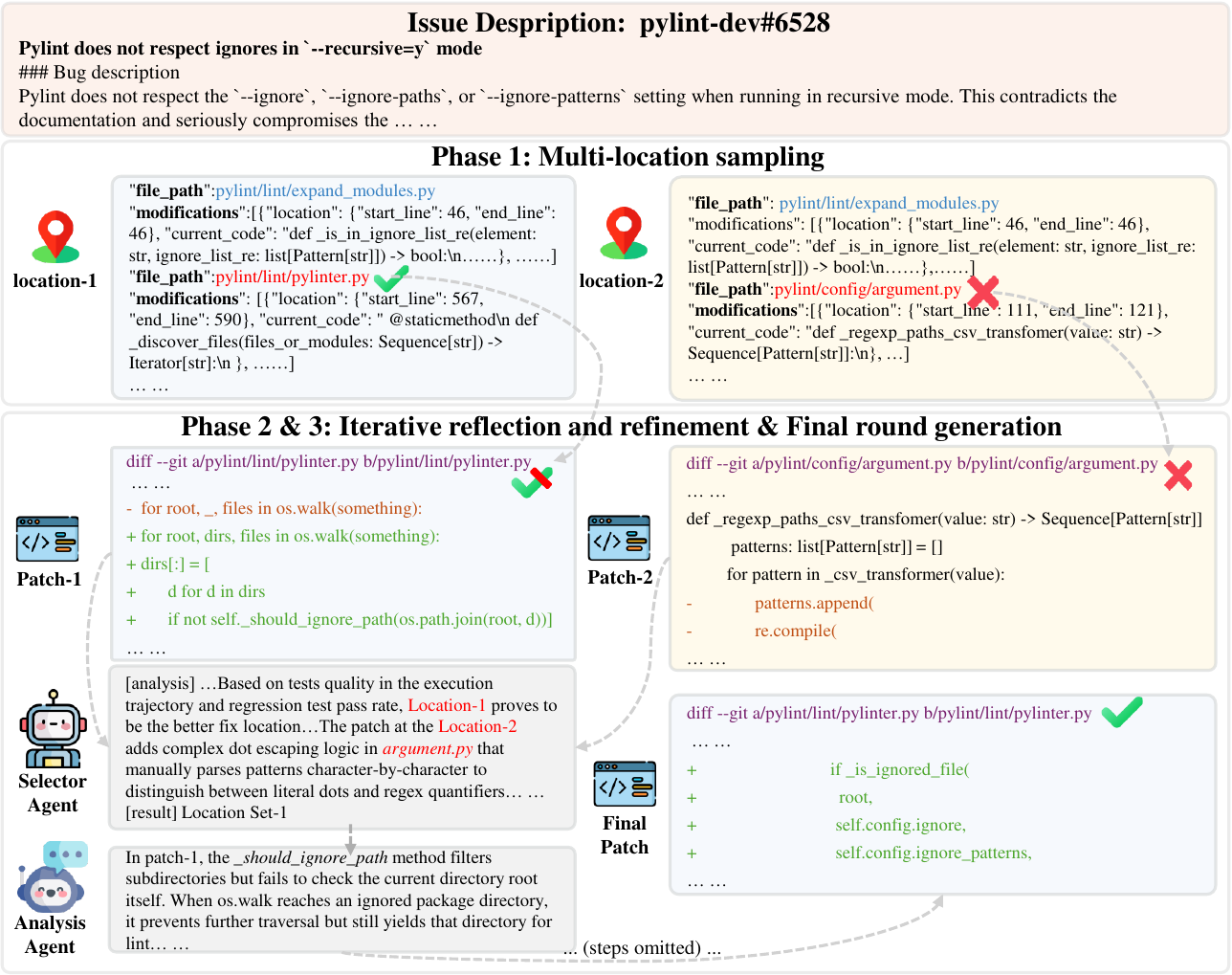}
    \caption{A case study of PhoenixRepair on pylint-dev\#6528.}
    \label{fig:casestudy}
\end{figure*}
\subsection{Case Study}

Figure~\ref{fig:casestudy} presents an example that demonstrates the effectiveness of our method in successfully resolving the issue \texttt{pylint-dev\#6528}.

In \textbf{Phase~1 Multi-location Sampling}, the location sampling agent performs iterative fault localization five times, resulting in five candidate edit locations. After deduplication, two distinct candidate edit locations are obtained.
Specifically, \emph{edit location-1} targets the \texttt{discover\_files} method in \texttt{pylint/lint/pylinter.py}, which functions as the main entry point for recursive file discovery during linting, and this location is correctly identified for modification. In contrast, \emph{edit location-2} focuses on the pattern processing logic in \texttt{pylint/config/argument.py}, which is responsible for handling ignore patterns specified via configuration arguments.

In \textbf{Phase~2 Iterative reflection and refinement}, the coder agent is constrained to generate patches independently for each of the identified edit locations. Then, we compress the execution trajectory associated with each patch by removing redundant parts from each trajectory. An analysis agent examines the compressed trajectory for each patch, focusing on the quality of reproduced tests and the edge tests. Meanwhile, we apply each patch to the original repository and perform regression testing to compute the regression pass rate for each patch.
Based on a joint assessment of tests quality and regression test pass rate, the selector agent determines that the patch generated at \emph{location-1} corresponds to a more promising fix location and thus selects it for further refinement. After selecting the final-round edit location, we proceed to Phase 3.

In \textbf{Phase~3 Final round generation}, an analysis agent examines the historical patches and execution trajectories generated at \textit{location-1}. The resulting analysis concludes that although Patch-1 correctly modifies the \texttt{discover\_files} method and introduces a \texttt{should\_ignore\_path} helper function to filter out subdirectories of the current directory, it fails to perform an essential check on the current directory \texttt{root} itself before executing the \texttt{yield} statement. As a result, directories that should be ignored are still yielded to the linting pipeline, even though traversal into their subdirectories is prevented. This oversight causes files within ignored package directories to be incorrectly included in subsequent linting steps.
Based on the actionable insights provided by the analysis agent, the coder agent addresses the identified deficiency by inserting an explicit \texttt{if \_is\_ignored\_file(root, \ldots)} check prior to yielding directory paths. This modification ensures that directories intended to be ignored are entirely excluded from recursive file discovery.

Consequently, the final patch correctly
prevents both traversal into ignored directories and the linting of their contained files, fully resolving the issue described in \texttt{pylint-dev\#6528}.

\section{Background and Related Work}

\revised{In the broader area of intelligent software engineering, large language models (LLMs) and software agents are increasingly used to automate complex development workflows~\cite{zheng2025towards,yang2025large,zheng2023survey,wang2026towards,wang2025agents,zhou2025adaptive,chen2024identifying,yang2024hyperion}. Recent studies have advanced a wide range of software engineering tasks, including code generation~\cite{shi2023sotana,li2024repomincoder,wang2025beyond,zheng2024top,zhang2025llm,li2025s,quan2025codeelo,si2025design2code,gu2025retrieve,chen2024rmcbench,zheng2024humanevo,wang2021code,wang2024rlcoder,lai2025analogcoder,zhu2025domaineval,nie2023unveiling,wang2026arkrepobench,wang2026realsec}, code search~\cite{gu2024secret,gu2025spencer,gong2025cosqa+,gu2022accelerating,hu2023revisiting,shi2023cocosoda,li2023rethinking,chen2023needs,wang2023you,hu2024tackling,dong2024improving,zheng2024costv,li2025search,chi2024empirical,wang2022enriching,chen2024decoder,zhang2023code}, automated issue resolution~\cite{guo2025omnigirl,tao2024magis,chen2025swe,ma2025alibaba,li2025swe,xie2025swe,chen2025prometheus,guo2026swe}, code summarization~\cite{shi2022evaluation,shi2021cast}, code translation~\cite{wang2024repotransbench,ou2024repository,pan2024lost,tao2024unraveling,yan2023codetransocean}, commit message generation~\cite{tao2022large,tao2024kadel,xue2024automated,zhang2024using,zhang2024automatic,tao2021evaluation,shi2022race,guo2023snippet,zhang2023ealink}, efficient model optimization~\cite{wang2024sparsecoder,guo2024stop,guo2023longcoder,gim2024prompt,cai2024pyramidkv,yue2024wkvquant,feng2024ada,wang2026logical}, and code understanding~\cite{bai2024longbench,wang2021cocosum,liao2025e2llm,li2025deepcircuitx,wang2026towards,zhou2025adaptive,wang2026reporeasoner,zhang2026conversation}. These efforts establish a solid methodological foundation for LLM-driven software agents. Automated issue resolution, the focus of this work, is one of the core tasks in this landscape, requiring agents to localize faults within real-world codebases and generate correct repair patches. Targeting this task, PhoenixRepair systematically expands the search space of repair strategies through multi-location sampling and an iterative reflection and refinement mechanism.}

\subsection{Fault Localization}

LLM-based fault localization has attracted increasing research attention~\cite{kang2024quantitative, wu2023large, xia2024agentless}.
For instance, Agentless~\cite{xia2024agentless} adopts a hierarchical localization powered by LLMs. Building on these foundations, agent-based methods enable automated codebase exploration through multi-step reasoning.
For example, 
OpenHands~\cite{wang2024openhands} utilizes bash commands like grep and file viewing tools. SWE-agent~\cite{yang2024swe} employs an Agent-Computer Interface for navigation.

Meanwhile, several studies have adopted graph-based representation methods to enhance code understanding by capturing dependencies between code files. Graph-based approaches typically construct dependency graphs that encode relationships among code entities, thereby enabling multi-hop reasoning across files. For example, RepoUnderstander~\cite{ma2024understand} constructs hierarchical graphs and function call graphs. GraphLocator~\cite{liu2025graphlocator} is an automated issue localization approach that constructs causal issue graphs (CIGs) to identify code locations requiring modification based on natural language issue descriptions.
OrcaLoca~\cite{yu2025orcaloca} adopts a simplified graph optimized with priority scheduling and context pruning, while LocAgent~\cite{chen2025locagent} leverages graph-based representations to enable multi-hop reasoning across code dependencies. Our localization approach leverages the epository navigation capabilities of agent-based methods while incorporating dependency information, thereby expanding the search space for fault localization.

\subsection{Automated Issue Resolution}
Automated software issue resolution aims to reduce the manual effort required to fix defects in real-world software projects. 
Recent research shows that by integrating these models with external tools and environmental interactions, LLMs can evolve into autonomous agents that address complex software engineering problems~\cite{qu2025tool, wang2024tools, wang2024trove, cai2023large}. 
SWE-search~\cite{antoniades2024swe} introduces a multi-agent system that employs MCTS.
SWE-Debate~\cite{li2025swe} presents an agent-based debate architecture designed to improve software issue resolution.
LingmaAgent~\cite{ma2024alibaba} condenses the repository into a knowledge graph and employs the MCTS strategy. OpenHands~\cite{wang2024openhands} supports code editing and execution through sandboxed environments and implements agents such as CodeActAgent for evaluating repairs.
SWE-agent~\cite{yang2024swe} leverages an Agent-Computer Interface to support repository navigation, file editing, and command execution for automated issue resolution.
Mini-SWE-agent~\cite{minisweagent2024} is a minimalist autonomous coding agent that maintains competitive performance. 
Live-SWE-agent~\cite{xia2025live} autonomously evolves its own agent scaffold on-the-fly during runtime, starting from a minimal bash-only baseline and self-improving while solving issues.
Meanwhile, procedure-based methods have also emerged, including Agentless~\cite{xia2024agentless}, which decomposes the task into localization, repair, and validation stages, and AutoCodeRover~\cite{zhang2024autocoderover}, which combines LLMs with sophisticated code search capabilities to generate program modifications.

Recent studies on SWE-bench have explored diverse directions in agent design, verification, and training.
InfCode~\cite{Li2025InfCodeAI} is an adversarial multi-agent framework for automated repository-level issue resolution.
SWE-Lego~\cite{Tao2026SWELegoPT} introduces a supervised fine-tuning (SFT) recipe aimed at achieving state-of-the-art performance on issue resolution.
RepoForge~\cite{Chen2025RepoForgeTA} is an autonomous, end-to-end pipeline that generates, evaluates, and trains SWE agents at scale.
Kimi-Dev~\cite{Yang2025KimiDevAT} introduces staged training as a pathway to more generalizable coding agents.
Self-play SWE-RL~\cite{Wei2025TowardTS} envisions an early step toward superintelligent software agents that amass extensive learning experience autonomously from repositories, ultimately surpassing human-level software understanding, improvement, and generation.

\subsection{Agent Enhancement Techniques}

Compared with traditional zero-shot or few-shot generation paradigms, language model agents that can interact with virtual environments~\cite{hong2023metagpt,sumers2023cognitive,wang2024survey} are becoming a mainstream research direction. They have shown strong performance in tasks such as 
web navigation~\cite{koh2024visualwebarena,yao2023webshop,yao2023react,zhou2023webarena} and code generation~\cite{wang2024rlcoder,jiang2026aligncoderaligningretrievaltarget,zhang2023repocoder}. At the same time, interactive decision making and code generation are becoming tightly integrated. Code is no longer used only as an output. Instead, it serves as the primary form of action for language model agents. It is used for tool construction, tool invocation, and the organization of complex reasoning processes. 

Recent research has developed various methods to enhance agent performance.
AutoGPT~\cite{yang2023auto} and AgentGPT~\cite{agentgpt2023}, integrate tool usage to extend the capabilities of agents.
Similarly, DeepAgent introduces an end-to-end deep reasoning framework that unifies autonomous reasoning, tool discovery, and action execution, while addressing the challenges of long-horizon interactions through a structured memory folding mechanism. Additionally, MemGPT~\cite{packer2023memgpt} enhances contextual understanding by incorporating memory mechanisms, enabling more efficient processing of information over time. 
These advancements in tool and memory usage contribute to the growing field of intelligent agents capable of handling complex, long-term tasks.

CodeMonkeys~\cite{ehrlich2025codemonkeys} is a problem-solving system designed specifically for scaling law compute. However, the inherent characteristics of probabilistic language models exhibit a strong bias toward high-probability solution patterns, thereby significantly restricting the diversity of the search space. For instance, the agent may confine its repair attempts to a single confident edit location throughout multiple sampling iterations.
SE-Agent~\cite{lin2025se} first samples multiple trajectories to construct a trajectory pool, and then incorporates mutation and recombination operations at the action steps within these trajectories. However, if all sampled trajectories focus on a specific high-confidence but incorrect edit location, even mutating and recombining the intermediate steps within these trajectories will not generate the correct fix.

\section{Threats to Validity}
In this section, we identify and discuss several potential threats to the validity of our study and the steps taken to mitigate them.

\textit{Internal Validity.} One potential threat arises from the implementation details of the compared approaches. To mitigate this threat, we strictly followed the experimental settings specified in each baseline's code repository and verified our implementations against the results reported by the original authors.

\textit{External Validity.}
One potential threat arises from the choice of backbone models used in our evaluation. Due to budget constraints, we primarily evaluate on five open-source or open-weight models. However, we mitigate this limitation by demonstrating consistent performance improvements across all five models, suggesting that the effectiveness of our framework is not tied to a specific model architecture. Furthermore, our generalizability experiments on Mini-SWE-agent and Live-SWE-agent show consistent trends, reinforcing the robustness of our findings.

\textit{Construct Validity.} One potential threat concerns whether the evaluation metrics used in our study are sufficient to fully capture the quality of automated issue resolution. We adopt Pass@1 as the primary evaluation metric, which assesses whether the first generated patch produced by the model successfully passes all associated tests for a given issue. In addition to this metric, we also report fault localization accuracy at multiple levels of granularity, including File, Module, and Function-level Acc@1. These metrics offer complementary perspectives and provide additional evidence of our framework's effectiveness in accurately identifying the correct edit locations.

Another potential threat concerns the evaluation of our iterative reflection and refinement mechanism. In particular, the assessment of the quality of reproduction tests and edge tests relies on the judgment of an analysis agent, which may introduce a subjective bias. To mitigate this threat, we do not rely solely on this qualitative assessment. Instead, we combine test quality evaluation with objective regression test pass rates. This design ensures that the patch selection process is informed by both qualitative judgments and quantitative performance indicators.

\section{Conclusion}
We present PhoenixRepair, a novel multi-agent framework that addresses the limitation of insufficient exploration in existing agent-based approaches for automated issue resolution. By systematically exploring multiple candidate edit locations and employing iterative reflection and refinement mechanisms, our framework effectively expands the search space of repair strategies. Our experimental results demonstrate that PhoenixRepair achieves state-of-the-art performance across five open-source backbone models on SWE-bench-Verified. Specifically, PhoenixRepair attains the highest absolute resolution rate of 76.0\% Pass@1 with MiniMax-M2.5, and delivers the largest relative improvement of 7.8\% over SWE-agent with DeepSeek-V3.1. In addition to overall resolution performance, PhoenixRepair also achieves higher fault localization accuracy across file, module, and function-level granularities, with particularly pronounced improvements at finer-grained levels such as the function level. Moreover, our two core mechanisms demonstrate strong generalizability, as they can be effectively transferred to other agent-based methods such as Mini-SWE-Agent and Live-SWE-Agent. This transferability validates the robust generalization capability of our core mechanisms.


\section*{Acknowledgments}
This work is supported by the National Natural Science Foundation of China (Grant No. 92582202, No. 62302534).

\bibliographystyle{IEEEtran}
\bibliography{sample-base}

\vfill

\end{document}